\documentclass[sigconf]{acmart}
\usepackage{multirow}
\usepackage{tikz}
\usetikzlibrary{tikzmark}
\usepackage{dashrule}
\usepackage{booktabs}
\usepackage{amsmath}
\usepackage{colortbl}
\usepackage[shortlabels]{enumitem}

\AtBeginDocument{%
  \providecommand\BibTeX{{%
    \normalfont B\kern-0.5em{\scshape i\kern-0.25em b}\kern-0.8em\TeX}}}

\renewcommand\footnotetextcopyrightpermission[1]{} 
\setcopyright{none}
\setlist[enumerate]{nosep}

\begin{document}

\title{Semi-supervised Text-based Person Search}


\author{Daming Gao}
\affiliation{
  \institution{Soochow University}
  \city{Suzhou}
  \country{China}
}
\email{dmgao@stu.suda.edu.cn}

\author{Yang Bai}
\affiliation{
  \institution{Soochow University}
  \city{Suzhou}
  \country{China}
}
\email{ybaibyougert@stu.suda.edu.cn}

\author{Min Cao}
\authornote{Corresponding author.}
\affiliation{
  \institution{Soochow University}
  \city{Suzhou}
  \country{China}
}
\email{mcao@suda.edu.cn}

\author{Hao Dou}
\affiliation{
  \institution{Meituan}
  \city{Beijing}
  \country{China}
}
\email{douhao05@meituan.com}

\author{Mang Ye}
\affiliation{
  \institution{Wuhan University}
  \city{Wuhan}
  \country{China}
}
\email{yemang@whu.edu.cn}

\author{Min Zhang}
\affiliation{
  \institution{Soochow University}
  \city{Suzhou}
  \country{China}
}
\email{minzhang@suda.edu.cn}


\renewcommand{\shortauthors}{Daming Gao et al.}
\begin{abstract}
Text-based person search (TBPS) aims to retrieve images of a specific person from a large image gallery based on a natural language description.
Existing methods rely on massive annotated image-text data to achieve satisfactory performance in fully-supervised learning.
It poses a significant challenge in practice, as acquiring person images from surveillance videos is relatively easy, while obtaining annotated texts is challenging.
The paper undertakes a pioneering initiative to explore TBPS under the semi-supervised setting, where only a limited number of person images are annotated with textual descriptions while the majority of images lack annotations.
We present a two-stage basic solution based on generation-then-retrieval for semi-supervised TBPS. 
The generation stage enriches annotated data by applying an image captioning model to generate pseudo-texts for unannotated images.
Later, the retrieval stage performs fully-supervised retrieval learning using the augmented data.
Significantly, considering the noise interference of the pseudo-texts on retrieval learning, we propose a noise-robust retrieval framework that enhances the ability of the retrieval model to handle noisy data. 
The framework integrates two key strategies: Hybrid Patch-Channel Masking (PC-Mask) to refine the model architecture, and Noise-Guided Progressive Training (NP-Train) to enhance the training process. 
PC-Mask performs masking on the input data at both the patch-level and the channel-level to prevent overfitting noisy supervision.
NP-Train introduces a progressive training schedule based on the noise level of pseudo-texts to facilitate noise-robust learning.
Extensive experiments on multiple TBPS benchmarks show that the proposed framework achieves promising performance under the semi-supervised setting.
\end{abstract}

\ccsdesc[500]{Computing methodologies~Artificial intelligence}
\ccsdesc[500]{Information systems~Image search}

\keywords{text-based person search; semi-supervised learning; noise-robust learning; learning with masking; curriculum learning}

\maketitle

\section{Introduction}
Text-based person search (TBPS)~\cite{li2017person,niu2024overview} aims at retrieving the target person images from a large image gallery using natural language descriptions, with promising applications in surveillance systems. 
TBPS shares connections with image-based person re-identification~\cite{ye2021deep, leng2019survey} and image-text retrieval~\cite{chen2021learning, ijcai2022-759} and yet presents unique challenges. Unlike image-based person re-identification, TBPS utilizes open-form textual queries for more flexible search information, while dealing with cross-modal discrepancies due to modality heterogeneity. Compared to image-text retrieval, TBPS focuses on person domain retrieval, introducing the challenge of smaller inter-class variation and requiring enhanced perception and reasoning over fine-grained information.

\begin{figure}[t]
\setlength{\abovecaptionskip}{0.2cm}
  \centering
  \includegraphics[width=0.99\linewidth]{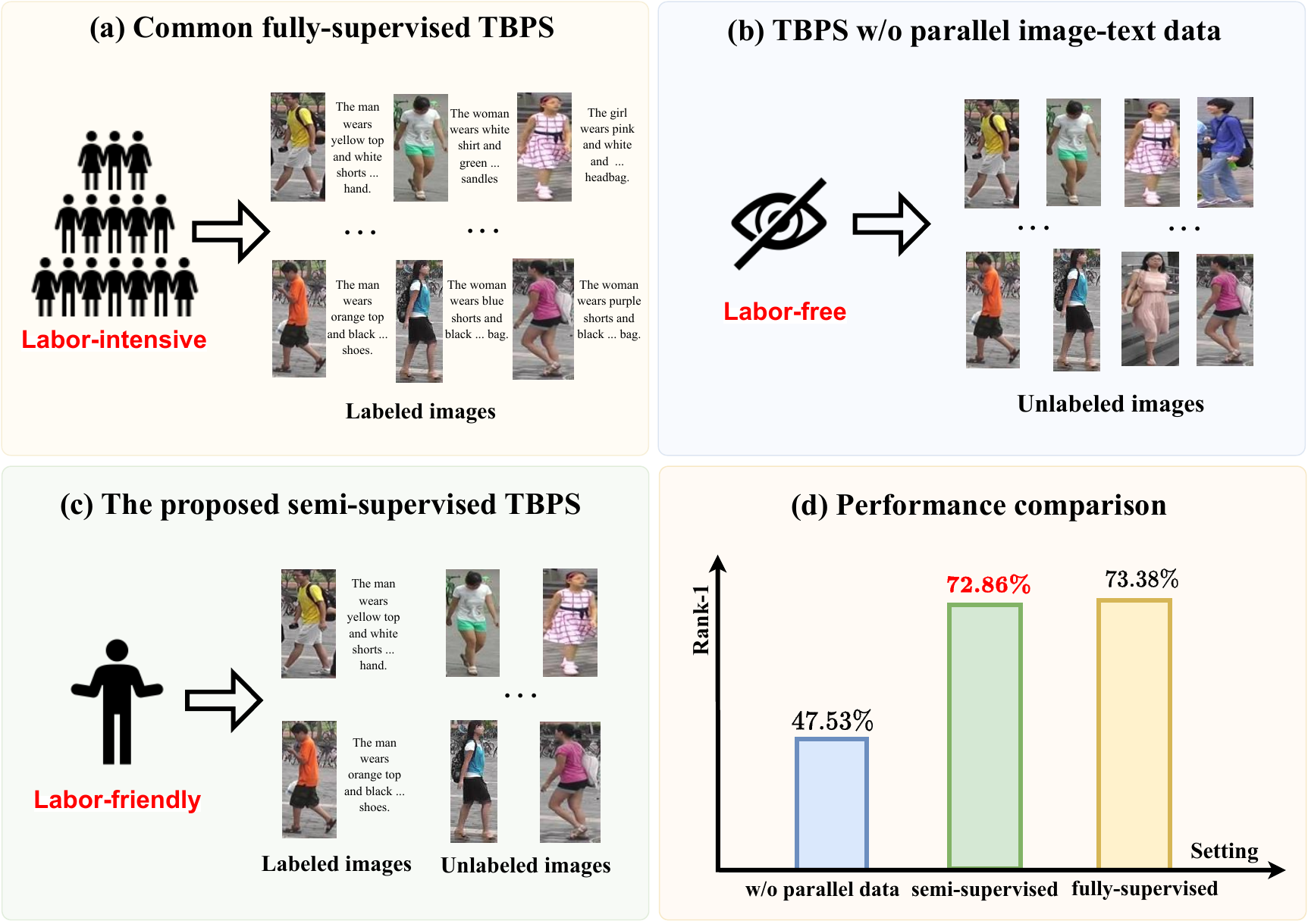}
  \caption{Comparisons with (a) common fully-supervised TBPS, (b) TBPS w/o parallel image-text data and (c) the proposed semi-supervised TBPS, and (d) performance comparison of the methods with above settings (i.e., the method~\cite{bai2023text} for TBPS w/o parallel image-text data and the method~\cite{jiang2023irra} for the fully-supervised TBPS) on CUHK-PEDES~\cite{li2017person}.}
  \label{fig:setting}
  \vspace{-0.55cm}
\end{figure}

To address these challenges, researchers have introduced various TBPS approaches~\cite{zhang2018deep, cao2023empirical, li2022learning, wang2020vitaa}, which focus on learning discriminative and modality-invariant feature representations, employing diverse fine-grained cross-modal alignment modules. 
These efforts have yielded promising performance improvements, and yet come at the cost of requiring a substantial number of image-text pairs for training the model (Fig.~\ref{fig:setting}(a)).
This requirement is impractical in real-world scenarios due to the manual annotation typically involved.
To address this limitation, few works~\cite{jing2020cross, zhao2021weakly, bai2023text} have explored TBPS without relying on a fully-supervised setting, thus reducing the burden of annotating.
Zhao et al.~\cite{zhao2021weakly} introduced a weakly supervised TBPS method that eliminates the need for identity labeling. 
Jing et al.~\cite{jing2020cross} developed a domain-adaptive TPBS approach, training a model on a source domain using supervised learning and then applying it to a new target domain. 
Bai et al.~\cite{bai2023text} were the pioneers of TBPS without parallel image-text data, removing the need for annotations.
The first two works did not completely eliminate the need for annotations in massive image-text pairs, whereas only the last one achieved a complete release of annotations (Fig.\ref{fig:setting}(b)). 
However, this work yields inadequate retrieval performance due to the complete absence of annotations (Fig.\ref{fig:setting}(d)).

Considering the trade-off between annotation cost and retrieval performance, this paper presents the first attempt at a practical setting: semi-supervised TBPS. In this setting, the model is trained with access to a small number of person image-text pairs and a large collection of person images~\footnote{In real-world scenarios, obtaining person images from surveillance videos via person detection technologies~\cite{yu2020scale, braun2019eurocity} is relatively straightforward, while acquiring paired textual descriptions often necessitates manual annotation.} (Fig.\ref{fig:setting}(c)).
While semi-supervised learning has been explored for general image-text retrieval~\cite{huang2024semi,kang2022intra}, its exploration within the TBPS community remains limited to the best of our knowledge.
Developing a semi-supervised TBPS solution is essential to achieve excellent performance with minimal annotation costs.
Nevertheless, TBPS poses unique challenges in semi-supervised learning compared to the general image-text retrieval task, primarily due to its specialization in the person domain and the inclined demand for modeling fine-grained information.

\begin{figure}[t]
\setlength{\abovecaptionskip}{0.1cm}
  \centering
  \includegraphics[width=0.96\linewidth, height=0.59\linewidth]{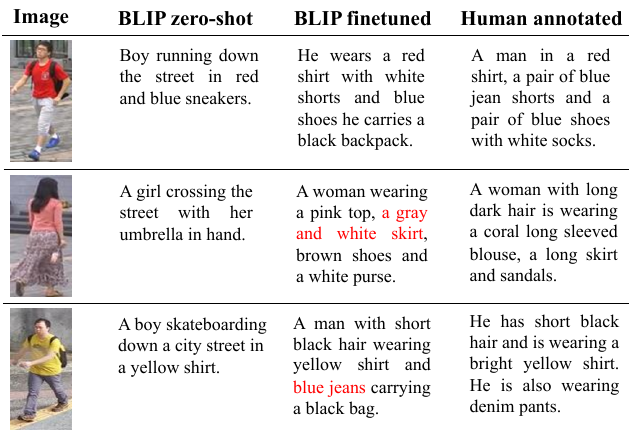}
  \caption{Visualization of human annotated texts and generated pseudo-texts from the vision-language model BLIP~\cite{li2022blip} under the zero-shot setting and finetuned on 1\% labeled data. Pseudo-texts from zero-shot BLIP tend to be coarse-grained while those from finetuned BLIP possess more fine-grained details but may contain inevitable noise. The noise is highlighted in red. More examples are shown in the Appendix.}
  \label{fig:text-visualization}
  \vspace{-0.4cm}
\end{figure}

For this, we firstly develop a two-stage basic solution based on generation-then-retrieval. In the generation stage, an off-the-shelf image captioning model finetuned on the few labeled data is applied to generate pseudo-textual descriptions for the unlabeled person images, and thus the deficiency of the labeled data can be alleviated by incorporating the pseudo-labeled data. In the retrieval stage, the retrieval model is trained with the combined few labeled data and pseudo-labeled data in a fully-supervised learning manner.
Remarkably, the fine-grained characteristic of TBPS brings the risk that the pseudo-texts may not always align well with the person images (BLIP finetuned \emph{vs.} Human annotated in Fig.~\ref{fig:text-visualization}). This inevitable noise may put the retrieval model at risk of learning misalignment.

To overcome this issue, we propose a noise-robust retrieval framework to enhance the ability of retrieval model to handle noisy labeled data.
This framework integrates noise-robust design principles to the model architecture and the training strategy. 
For the model architecture, we introduce a hybrid patch-channel masking (PC-Mask), combining patch-level and channel-level masking.
Patch-level masking randomly masks a portion of the input data in the original semantic space, while channel-level masking randomly masks feature values in the computed representation space.
By leveraging their complementary effects, PC-Mask effectively decouples the underlying noisy correspondence in image-text pairs and prevents the retrieval model from overfitting the noisy supervision.
For the training strategy, we design a noise-guided progressive training (NP-Train), inspired by curriculum learning~\cite{wang2021curriculumLearningSurvey}. 
NP-Train schedules the training in a progressive manner. It starts by utilizing more reliable data with less noise and gradually introduces more challenging data with higher noise levels. This strategy ensures that the model is dominated by high-confidence data, alleviating the interference from noisy data. 
By applying PC-Mask and NP-Train to the retrieval model (e.g., RaSa~\cite{bai2023rasa} and IRRA~\cite{jiang2023irra}), we establish a noise-robust retrieval framework, which facilitates the cross-modal representation learning in the presence of noisy training data.

Our contributions can be summarized as follows.
(1) We make the first exploration of semi-supervised TBPS, which is a more resource-friendly and realistic scenario than the much-studied fully-supervised TBPS in the current community. A two-stage solution based on generation-then-retrieval is presented to address this novel setting.
(2) In order to tackle the noise interference resulting from generated pseudo-labeled data during the training of retrieval model, we propose a noise-robust retrieval framework. It integrates two key strategies: hybrid patch-channel masking for refining the model architecture, and noise-guided progressive training for enhancing the training process. 
(3) Extensive experiments demonstrate that the proposed framework can achieve promising performance under the semi-supervised setting.

\section{Related Work}
\subsection{Text-based Person Search}
Current TBPS methods~\cite{zhang2018deep, bai2023rasa, cao2023empirical, li2022learning,wang2020vitaa,jiang2023irra} can be classified into two categories: cross-modal alignment and pretext task design. The former aligns visual and textual features in a shared embedding space, while the latter designs pretext tasks for learning modality-invariant features.
Specifically, for cross-modal alignment, early studies focused on global alignment~\cite{zhang2018deep, zheng2020dual} but later advanced to local alignment~\cite{chen2022tipcb, niu2020improving}, such as patch-word or region-phrase correspondences. 
Also, self-adaptive methods~\cite{gao2021contextual, li2022learning} were proposed to learn multi-granularity alignment, and some studies~\cite{wang2020vitaa, jing2020pose} incorporated external technologies like human parsing and pose estimation to aid alignment.
For pretext task design, Wu \emph{et al.}~\cite{wu2021lapscore} designed two color-reasoning sub-tasks to enable models to learn representations; Bai \emph{et al.}~\cite{bai2023rasa} formulated the relation and sensitivity-aware learning tasks to address weak cross-modal correspondence; Jiang \emph{et al.}~\cite{jiang2023irra} introduced a cross-modal implicit relation reasoning task to enhance the learning of fine-grained representation.

The mentioned methods heavily depend on fully annotated data, which is challenging to obtain in real-world scenarios due to its cost and time requirements. Several works~\cite{jing2020cross, zhao2021weakly, bai2023text} have begun exploring TBPS in non-supervised settings, aiming to alleviate the burden of annotation from various perspectives.
Jing \emph{et al.}~\cite{jing2020cross} addressed the domain-adaptive TPBS, where they adapted the proposed moment alignment model trained on a source domain to a new and different target domain.
Zhao \emph{et al.}~\cite{zhao2021weakly} introduced weakly supervised TBPS, eliminating the need for identity labeling. They employed a mutual training framework to generate and refine pseudo labels.
Bai \emph{et al.}~\cite{bai2023text} pioneered TBPS without parallel image-text data, completely removing the requirement for textual annotations. They explored a fine-grained image captioning strategy to generate pseudo-texts and a confidence-based training scheme to ensure reliable representation learning.
However, the domain-adaptive TPBS~\cite{jing2020cross} requires abundant annotated data in the source domain, while weakly supervised TBPS~\cite{zhao2021weakly} still requires textual annotations for a large number of images, making both approaches impractical due to the expense and time involved in annotation. Although Bai \emph{et al.}~\cite{bai2023text}  eliminated the need for textual annotation, generating pseudo-texts for the task requires additional prior knowledge and intricate rule design. This leads to lower-quality generated texts and limited performance in training retrieval models.

To this end, this paper explores semi-supervised TBPS, aiming to annotate a small amount of data while maximizing the use of readily available resources, specifically person images from surveillance cameras, to achieve satisfactory performance.
Both Bai \emph{et al.}~\cite{bai2023text} and our work adopt the generation-then-retrieval paradigm, where pseudo-labeled data is initially generated and used for retrieval training. However, our work focuses on noise-robust learning with pseudo-labeled data in the retrieval stage, while Bai \emph{et al.}~\cite{bai2023text} emphasizes the image captioning strategy in the generation stage.

\subsection{Learning with Noisy Data}
The issue of noisy data in cross-modal scenarios has gained increasing attention recently, with a focus on learning robust representations from mismatched multimodal data pairs. Several methods have been proposed to mitigate the negative impact of noisy supervision.
For example, Huang \emph{et al.}~\cite{huang2021ncr} introduced the NCR framework, which partitions data into clean and noisy subsets based on neural network memorization and rectifies correspondence adaptively. 
Qin \emph{et al.}~\cite{qin2022decl} developed an evidential learning paradigm that captures noise uncertainty and mitigates its adverse effects with a dynamic hinge loss. 
Han \emph{et al.}~\cite{han2023meta} approached the problem from a meta-learning standpoint and proposed a data purification strategy to remove noisy samples. 
Additionally, Qin \emph{et al.}~\cite{qin2023nc-tbps} addressed the noisy correspondence issue in TBPS, employing a confident consensus mechanism to partition noisy samples and introducing triplet alignment loss to reduce the misleading risks in training. 
The previous studies address the noisy data problem in completely mismatched image-text pairs. However, in our scenario, the pseudo-texts are weakly correlated with the person images, containing partial semantic noise interference. Thus, these existing approaches cannot be directly applied. To achieve robust learning with weakly labeled pseudo-texts, we propose NP-Train, which schedules training based on quantified noise scores, and PC-Mask, which provides additional learning regularization.

\subsection{Learning with Masking}
In natural language processing, masked language modeling (MLM) derived from BERT~\cite{devlin2018bert} is a widely used pretraining task, which masks parts of input sentences and trains models to predict the missing content, resulting in pretrained representations with strong generalization capabilities~\cite{wang2018glue, rajpurkar2016squad,zellers2018swag}.
Similarly, the masked autoencoder (MAE)~\cite{he2022mae} applies the concept to computer vision by masking and reconstructing random image patches, leading to generalized representations and superior performance.
Inspired by MAE, FLIP~\cite{li2023flip} leverages masking for vision-language pretraining with expanded scale.
The proposed PC-Mask follows a similar paradigm but differs in its purpose. Instead of reconstruction, we utilize masking as a regularization strategy to address the noisy interference from pseudo-labeled training data.

\subsection{Curriculum Learning}
Curriculum learning (CL), introduced by Bengio \emph{et al.}~\cite{bengio2009curriculumLearning}, is a training strategy that starts with easy examples and progresses to more difficult ones, resembling human learning curricula. CL is commonly used for denoising by prioritizing high-confidence, less complex data to mitigate the impact of noise~\cite{wang2021curriculumLearningSurvey}.
For instance, Jiang \emph{et al.}~\cite{jiang2018mentornet} proposed MentorNet, which applies a data-driven curriculum to focus on potentially correct-labeled data, significantly improving network generalizability on corrupted data.
Guo \emph{et al.}~\cite{Guo2018curriculumnet} designed a curriculum that quantifies noise levels using cluster density measurements, enabling robust training on noisy web images.
Inspired by CL for denoising, we propose NP-Train to learn reliable cross-modal relations on noisy data.

\begin{figure*}
\setlength{\abovecaptionskip}{0.2cm}
  \centering
  \includegraphics[width=0.97\textwidth]{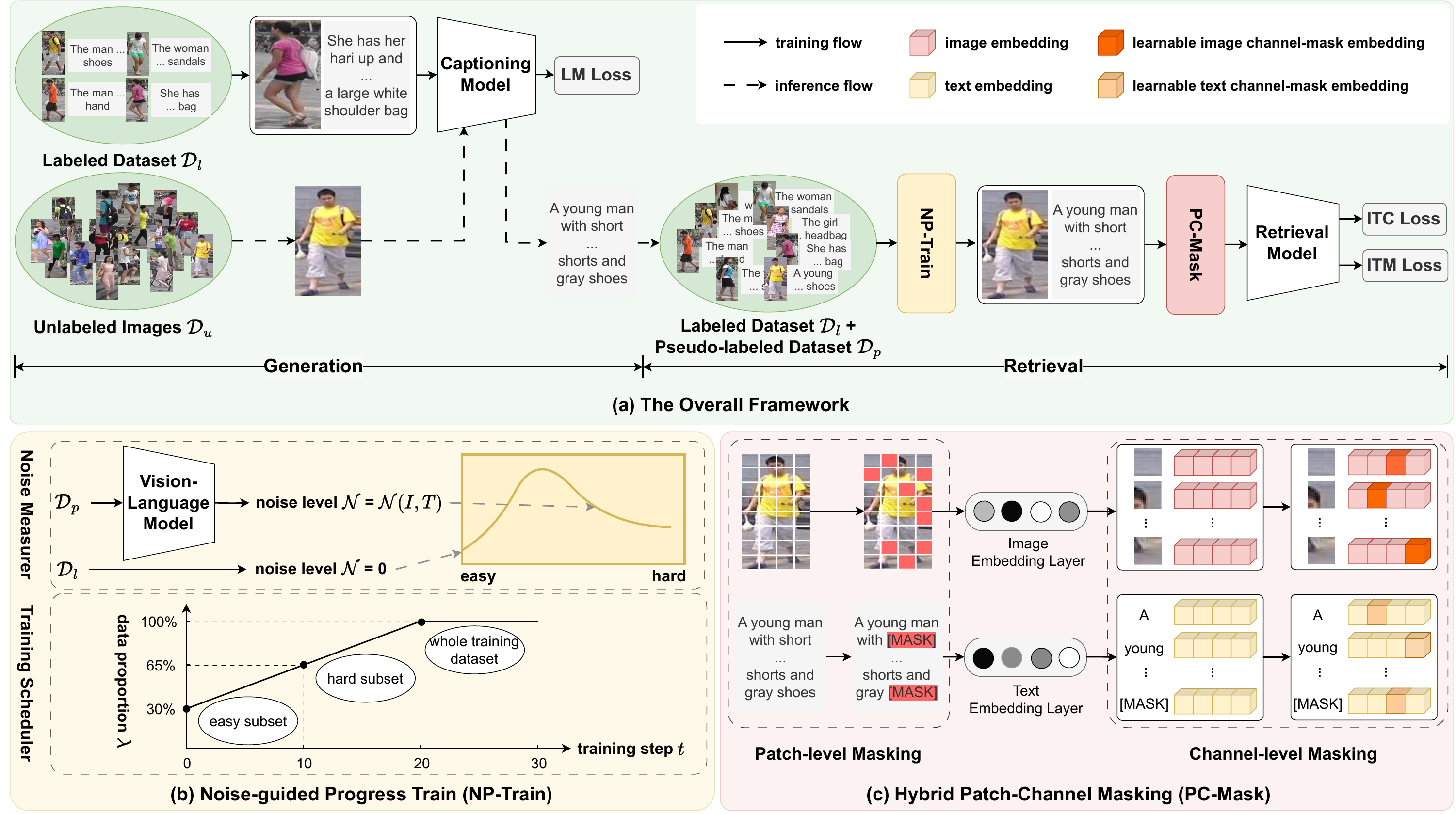}
  \caption{Illustration of the proposed generation-then-retrieval solution along with the noise-robust retrieval framework.}
  \label{fig:arch}
\vspace{-0.25cm}
\end{figure*}

\section{Methodology}
In semi-supervised text-based person search, we are given a labeled dataset $\mathcal{D}_l = \{I_i, T_i\}_{i=1}^{N_l}$. Here, $I_i$ denotes the $i$-th image, $T_i$ is the $i$-th textual description paired with $I_i$, $N_l$ is the total number of image-text pairs. Notably, this number is significantly smaller compared to the fully supervised counterpart.
Additionally, we are also provided with an unlabeled dataset of person images $\mathcal{D}_u = \{I_i\}_{i=1}^{N_u}$, wherein corresponding text annotations are absent.
Building upon the generation-then-retrieval solution, we introduce a noise-robust retrieval framework that aims to achieve promising performance utilizing both labeled data ($\mathcal{D}_l$) and unlabeled data ($\mathcal{D}_u$).
Fig.~\ref{fig:arch} provides an overview of the proposed generation-then-retrieval solution, which incorporates a noise-robust retrieval framework.

In the subsequent sections, we begin by outlining the basic two-stage solution rooted in the generation-then-retrieval course. We then detail the noise-robust retrieval framework, including the proposed hybrid patch-channel masking and noise-guided progressive training. For simplicity, we will use $I$ and $T$ to represent an image and a text, respectively, while omitting the symbolic subscripts.

\subsection{Basic Solution: Generation-Then-Retrieval}
\label{section:GTR}
To handle the semi-supervised TBPS, we firstly introduce a basic two-stage solution. 
In the generation stage, we leverage the few labeled data ($\mathcal{D}_l$) to generate pseudo-labels for the unlabeled data ($\mathcal{D}_u$), to enrich the training corpus.
In the retrieval stage, a retrieval model is trained, utilizing both the labeled data and the generated pseudo-labeled data in a supervised manner.

\subsubsection{\textbf{Generation.}} 
\hfill \\
Generating textual descriptions for person images in dataset $\mathcal{D}_u$ can be viewed as an image captioning task~\cite{xu2015show}, for which an off-the-shelf image captioning model can be employed.
However, it is important to note that current image captioning models~\cite{li2022blip, li2023blip2} are typically trained on generic image-text corpus data, which results in a noticeable gap when dealing with person-specific data. 
If directly applying such models to person images, the generated texts may lack fine-grained details and fail to capture the subtle information related to the appearance of individuals, as shown in Fig.~\ref{fig:text-visualization} (BLIP zero-shot \emph{vs.} Human annotated). Consequently, this can harm the subsequent retrieval stage, which heavily relies on precise cross-modal alignment learning.

Hence, we employ the few labeled data $\mathcal{D}_l$ to finetune the image captioning model, so that the person-related knowledge and the fine-grained captioning ability of the model can be learned. 
Specifically, given an image $I$ and its paired text $T = \{u_1, u_2, \cdots, u_n\}$ from $\mathcal{D}_l$, where $u_i$ is the $i$-th token in the text, we minimize the following language modeling loss during finetuning:
\begin{equation}
    \mathcal{L}_{lm} = - \sum_{i=1}^{n} \mathrm{log}\, P(u_i|I, u_1, u_2, \cdots, u_{i-1};\Theta),
\end{equation}
where the conditional probability $P$ is computed using an image captioning model with parameters $\Theta$.

After finetuning on the few labeled data, we use the image captioning model to generate pseudo-texts for each unlabeled person image within dataset $\mathcal{D}_u$. The resulting collection of pseudo-labeled data is represented as $\mathcal{D}_p = \{I_i, T_i\}_{i=1}^{N_p}$, where $N_p$ is the total number of pseudo-labeled image-text pairs.

\subsubsection{\textbf{Retrieval.}}
\label{Two-stage Solution: Retrieval}
\hfill \\
We combine the few pre-existing labeled dataset $\mathcal{D}_l$ with the newly-generated pseudo-labeled dataset $\mathcal{D}_p$ to create the training set for the retrieval stage. The default retrieval model, BLIP~\cite{li2022blip}, is adopted, and its network architecture, described here, serves as the foundation for the proposed noise-robust retrieval framework. 
BLIP is an advanced vision-language model with remarkable performance on various vision-language tasks. It has three modules.

\textbf{Image Encoder.} 
Visual transformer~\cite{dosovitskiy2020image} is employed as the image encoder.
Given an input image $I$, we first splits it into a sequence of $M$ non-overlapping patches, $p_i \in \mathbb{R}^{h \times w}$ ($i = 1, \cdots, M$).
Then a linear projection $\textbf{L}^v$ operates on the patches and obtain 1D tokens $v_i \in \mathbb{R}^{d}$.
It can be formulated as:
\begin{equation}
    \mathcal{V} = [v_{cls}, \textbf{L}^v(p_1), \textbf{L}^v(p_2), \cdots, \textbf{L}^v(p_M)] + P^v,
    \label{equation:ie}
\end{equation}
where $v_{cls} \in \mathbb{R}^{d}$ is a learnable classification token and $P^v \in \mathbb{R}^{(M+1) \times d}$ is the positional embedding.

Next, we send the embedding sequence $\mathcal{V}$ into the image encoder $\textbf{E}^{v}$ to extract image features $\mathcal{F}_v \in \mathbb{R}^{(M+1) \times d}$, which can be formulated as:
\begin{equation}
    \mathcal{F}^v = \textbf{E}^{v}(\mathcal{V}),
    \label{equation:ie1}
\end{equation}
wherein $f^v_{cls}\in \mathbb{R}^{d}$ represents the global feature of the image $I$.

\textbf{Text Encoder.} 
It shares the same architecture with BERT~\cite{devlin2018bert}. Given a input text $T$, we first tokenizes it into a sequence of $N$ tokens $w_i$ ($i = 1, \cdots, N$).
Then a embedding layer $\textbf{L}^t$ is employed to process the tokens and generate 1D embeddings $t_i \in \mathbb{R}^{d}$. This process can be expressed as follows:
\begin{equation}
    \mathcal{T} = [t_{cls}, \textbf{L}^t(w_1), \textbf{L}^t(w_2), \cdots, \textbf{L}^t(w_N)] + P^t,
    \label{equation:te}
\end{equation}
where $t_{cls} \in \mathbb{R}^{d}$ is a learnable embedding and $P^t \in \mathbb{R}^{(N+1) \times d}$ is the positional embedding of text data.

The embedding sequence $\mathcal{T}$ is passed via the text encoder $\textbf{E}^{t}$ to extract text features $\mathcal{F}^t \in \mathbb{R}^{(N+1) \times d}$. It is formulated as follows:
\begin{equation}
    \mathcal{F}^t = \textbf{E}^{t}(\mathcal{T}),
    \label{equation:te1}
\end{equation}
wherein $f^t_{cls}\in \mathbb{R}^{d}$ is the global feature of the text $T$.

\textbf{Image-grounded Text Encoder.} 
The objective is to integrate the image and text features, enabling comprehensive interaction between the visual and textual modalities. 
The encoder enhances the text encoder by incorporating an extra cross-attention layer within each transformer block.
Specifically, by utilizing the text embeddings $\mathcal{T}$ as query and the image features  $\mathcal{F}^v$ as key and value, the encoder $\textbf{E}^{m}$ generates a sequence of multimodal fused features $\mathcal{F}^m \in \mathbb{R}^{(N+1) \times d}$, which can be formulated as:
\begin{equation}
    \mathcal{F}^m = \textbf{E}^{m}(\mathcal{T}, \mathcal{F}^v),
    \label{equation:ite}
\end{equation}
wherein $f^m_{cls}$ is the fused feature of the image $I$ and the text $T$.

\textbf{Training Objectives.} 
Following BLIP, we optimize the retrieval model using the image-text contrastive loss (ITC) and image-text matching loss (ITM).
ITC aligns the feature space of the image encoder and text encoder by bringing positive image-text pairs closer and pushing negative pairs apart, and operates on $f^v_{cls}$ and $f^t_{cls}$.
ITM aims to learn the multimodal fused feature $f^m_{cls}$ that captures cross-modal alignment by predicting the positivity (match) or negativity (mismatch) of image-text pairs.
Refer to the original paper of BLIP~\cite{li2022blip} for the formula details.

The overall training objective is defined as:
\begin{equation}
    \mathcal{L} = \mathcal{L}_{itc} + \mathcal{L}_{itm} .
    \label{equation:ot}
\end{equation}
We train the retrieval model BLIP with Eq.~\ref{equation:ot} on the training sets $\mathcal{D}_p$ and $\mathcal{D}_L$ in a supervised manner.

\subsection{Hybrid Patch-channel Masking}
\label{section:PC-Mask}
As the pseudo-texts in $\mathcal{D}_p$ contain noise that can hinder the retrieval model's ability to learn precise cross-modal alignment, we enhance the above basic solution by integrating Hybrid Patch-channel Masking (PC-Mask) into the retrieval model for alleviating the impact of noise. PC-Mask applies masking at both the patch-level and the channel-level, as introduced in the following section.

\subsubsection{\textbf{Patch-level Masking.}}
\hfill \\
This strategy masks parts of the input data in the original semantic space, which is conducted on both the visual and textual modality. 

For the visual modality, given the sequence of patches from an input image $I$, we randomly mask out a portion of the patches with the masking ratio $\rho ^v$ by simply removing them and the remaining patches are inputted into the network. 
Hence, Eq.~\ref{equation:ie} is rewritten as:
\begin{equation}
    \mathcal{V}' = [v_{cls}, \textbf{L}^v(p_1), \textbf{L}^v(p_2), \cdots, \textbf{L}^v(p_{M'})] + P^v,
    \label{equation:pmv}
\end{equation}
where $M'=M(1-\rho ^v)$ is the number of unmasked patches.

In terms of the textual modality, given the sequence of tokens from an input text $T$, we randomly sample a subset of the tokens with the masking ratio $\rho ^t$ and replace them with the special [MASK] token. The masked sequence of tokens are passed to the network and thus Eq.~\ref{equation:te} is rewritten as:
\begin{equation}
    \mathcal{T}' = [t_{cls}, \textbf{L}^t(w_1), \textbf{L}^t([\mathrm{MASK}]), \cdots, \textbf{L}^t(w_N)] + P^t,
    \label{equation:pmt}
\end{equation}

\subsubsection{\textbf{Channel-level Masking.}}
\hfill \\
The strategy involves masking the feature channels within the computed representation space and is also employed in both the visual and textual modalities.

For the visual modality, we mask a subset of channel values in the patch embeddings $\mathcal{V}'$ in Eq.~\ref{equation:pmv}.
Specifically, we first incorporate a learnable channel-mask embedding $c^v \in \mathbb{R}^{d}$ as a prototype.
Next, for each embedding $v_i$ ($i=1,2, \cdots, M'$) in Eq.~\ref{equation:pmv}, we randomly sample a portion of its channel values using the masking ratio $\beta ^v$. These sampled values are then replaced with the corresponding channel values from $c^v$. This process can be represented as:
\begin{equation}
\mathcal{V}'' = [v_{cls}, \textbf{M}(\textbf{L}^v(p_1)), \textbf{M}(\textbf{L}^v(p_2)), \cdots, \textbf{M}(\textbf{L}^v(p_{M'}))] + P^v,
\end{equation}
\begin{equation} 
\textbf{M}(\textbf{L}^v(p_i)) = \textbf{L}^v(p_i) \odot b^v + c^v \odot (1 - b^v), 
\quad i=1 \ldots M' 
\end{equation}
where $b^v \in \mathbb{R}^{d}$ is a random binary mask vector (0 with probability $\beta ^v$ and 1 with probability $1-\beta ^v$ ), and $\odot$ is the hadamard product.

The sequence of embeddings $\mathcal{V}''$ is then sent to the image encoder to obtain the final features $\mathcal{F}_v''$, as depicted in Eq.~\ref{equation:ie1}.

Similar to the operations performed on the visual modality, we apply channel-level masking to the embedding sequence $\mathcal{T}'$ in Eq.~\ref{equation:pmt} for the textual modality. This masking involves a learnable channel-mask embedding $c^t \in \mathbb{R}^{d}$ and the masking ratio $\beta^t$. 
The outputted masked embeddings $\mathcal{T}''$ are then fed into the text encoder to obtain the final features $\mathcal{F}_t''$, as shown in Eq.~\ref{equation:te1}.

Using the masked textual embeddings $\mathcal{T}''$ and the masked visual features $\mathcal{F}_v''$, we generate the multimodal fused features $\mathcal{F}_m''$ by inputting them into the image-grounded text encoder as described in Eq.~\ref{equation:ite}.
During the training process, we optimize the ITC and ITM loss functions using the masked features $\mathcal{F}_v''$, $\mathcal{F}_t''$, and $\mathcal{F}_m''$. During inference, we apply the retrieval model to the original image and text inputs without masking.

In short, PC-Mask decouples the noisy correspondence across modalities by randomly masking data. This random operation can be considered as a form of data regularization to prevent overfitting during training with noisy data.

\subsection{\textbf{Noise-guided Progressive Training}}
\label{section:NP-Train}

In addition to PC-Mask, we also developed a Noise-Guided Progressive Training strategy (NP-Train) aimed at achieving noise-robust training. Instead of treating all generated pseudo-labeled data equally during training, regardless of their varying levels of noise, NP-Train follows a progression from more reliable data with less noise to more challenging data with higher levels of noise.

In this process, two modules are designed: the \textit{Noise Measurer}, which evaluates the noise level of each pseudo-text, and the \textit{Training Scheduler}, which determines the sequence of data subsets throughout the training process based on the noise level of data. In the following, we elaborate on these two modules.

\subsubsection{\textbf{Noise Measurer}}
\hfill \\
The generated pseudo-texts exhibit varying degrees of noise resulting from errors in the image captioning model. This noise pertains to descriptions semantically inconsistent or unrelated to the corresponding images, impeding the model's ability to learn accurate cross-modal alignment.

In this context, the noise measurer assesses the level of noise in each pseudo-text by considering the cross-modal semantic similarity. When a pseudo-text closely aligns with the depicted image, it indicates the strong alignment of cross-modal semantics, resulting in a lower potential for noise in the pseudo-text. Conversely, if a pseudo-text significantly differs from its corresponding image, it signifies a large semantic gap between the image-text pair, suggesting a higher likelihood of noise in the pseudo-text.

\begin{table*}
\setlength{\abovecaptionskip}{0.1cm}
    \centering
    \caption{Ablation studies on each component of the proposed framework on the widely-used datasets with 1\% labeled data.}
    \begin{tabular}{l|c|cc|c|cc|cc|cc}
    \toprule
    \multirow{3}{*}{\textbf{Methods}} & \multicolumn{4}{c|}{\textbf{Components}} & \multicolumn{2}{c|}{\textbf{CUHK-PEDES}} & \multicolumn{2}{c|}{\textbf{ICFG-PEDES}} & \multicolumn{2}{c}{\textbf{RSTPReid}} \\
    \cline{2-11}
     & \multirow{2}{*}{\textbf{Basic-GTR}} & \multicolumn{2}{c|}{\textbf{PC-Mask}} & \multirow{2}{*}{\textbf{NP-Train}} & \multirow{2}{*}{\textbf{R-1}} & \multirow{2}{*}{\textbf{mAP}} & \multirow{2}{*}{\textbf{R-1}} & \multirow{2}{*}{\textbf{mAP}} & \multirow{2}{*}{\textbf{R-1}} & \multirow{2}{*}{\textbf{mAP}} \\
     \cline{3-4}
     & & \textbf{P-Mask} & \textbf{C-Mask} & & & & & & & \\
    \midrule
    Baseline & - & - & - & - & 54.91 & 47.61 & 37.14 & 16.70 & 52.60 & 37.78  \\
    \hline
    M1 & \checkmark & - & - & - & 61.94 & 55.59 & 44.30 & 25.12 & 55.50 & 42.40 \\
    M2 & \checkmark & \checkmark & - & - & 63.01 & 56.64 & 45.63 & 26.29 & 55.40 & 43.43 \\
    M3 & \checkmark & - & \checkmark & - & 62.48 & 56.02 & 45.33 & 25.46 & 56.15 & 43.79 \\
    M4 & \checkmark & \checkmark & \checkmark & - & 63.35 & 57.06 & 45.71 & 26.32 & \textbf{56.55} & 43.57 \\
    M5 & \checkmark & - & - & \checkmark & 63.06 & 56.38 & 46.14 & 26.64 & 55.20 & 42.97 \\
    \hline
    Ours & \checkmark & \checkmark & \checkmark & \checkmark & \textbf{63.87} & \textbf{57.18} & \textbf{46.46} & \textbf{26.90} & 56.45 & \textbf{44.45} \\
    \bottomrule
    \end{tabular}
    \label{tab:Effectiveness-of-Each-Component}
\vspace{-0.1cm}
\end{table*}
Formally, given a pseudo-text $T$ and its corresponding image $I$ from the pseudo-labeled training set $\mathcal{D}_p$, the level of noise in $T$ is defined to have a negative correlation with their cross-modal similarity, as expressed below:
\begin{equation}
    \mathcal{N}(T) = 1 - \psi(I, T),
    \label{enquation:em}
\end{equation}
where $\psi(I, T)$ denotes the matching score of the pair and is computed by the state-of-the-art vision-language models. These models are typically pretrained on massive image-text data, enabling them to capture the cross-modal semantic relationship accurately.
By default, we employ the BLIP~\cite{li2022blip} to calculate the matching scores. 
Other variants are also used and compared for a comprehensively analysis of noise measurers in Table~\ref{tab:Comparison-between-Different-Noise-Measurers}.

\subsubsection{\textbf{Training Scheduler}} 
\hfill \\
Using the proposed noise measurer, we arrange the training image-text pairs in $\mathcal{D}_l$ + $\mathcal{D}_p$ in ascending order based on their assessed noise levels\footnote{The noise level of manually annotated texts in $\mathcal{D}_l$ is assigned as zero.}.
We schedule the training process by sequentially using these organized training data to optimize the retrieval model. Specifically, we employ the linear scheduler, which linearly adjusts the training data subset at each epoch. It can be formulated as:
\begin{equation}
    \lambda_{\text {linear}}(t)=
    \min \left(1, \lambda_0+\frac{1-\lambda_0}{T_{\text {grow}}} \cdot t\right),
\end{equation}
where $\lambda_0$ is the initial ratio of the easiest available samples, $T_{\text {grow}}$ is the epoch when all training samples are used for the first time and $t$ is the current training epoch number.
The function $\lambda_{\text {linear}}(t)$ is a linear function that maps the training epoch number to a scalar $\lambda \in (0,1]$. It determines the ratio of the easiest training data available at the $t$-th epoch. This function is monotonically non-decreasing, starting at $\lambda(0)>0$ and concluding at $\lambda(T) = 1$.
Referring to the method~\cite{wang2021curriculumLearningSurvey}, $\lambda_0$ is set to $0.3$ and $T_{\text {grow}}$ is set to $15$ in our experiments.

\section{Experiments}
We conduct experiments on three widely-used TBPS datasets: CUHK-PEDES~\cite{li2017person}, ICFG-PEDES~\cite{ding2021semantically} and RSTPReid~\cite{zhu2021dssl}, and a newly released dataset: UFine6926~\cite{zuo2023ufinebench}. 
For semi-supervised TBPS, we randomly select a small ratio\footnote{Experiments under higher ratios of labeled data are shown in the Appendix.} (1\%, 5\%, and 20\%) of labeled training data consisting of person images and their corresponding textual descriptions from the original benchmark set. The remaining person images, without their accompanying texts, are used as the unlabeled dataset in our experiments.
We adopt the Rank-k (R-k for simplicity, k=$1, 5, 10$) as the main evaluation metric and the mean average precision (mAP) as a complementary metric. 
\textit{The introduction of each dataset, evaluation metrics and the implementation details of the proposed framework are shown in the Appendix.}

\subsection{Backbones}
The proposed solution offers flexibility in selecting image captioning and text-based person retrieval backbones. 
To verify the scalability of the proposed framework, we employ two backbones for captioning in the generation stage: BLIP~\cite{li2022blip} and {BLIP-2}~\cite{li2023blip2}, and three retrieval backbones in the retrieval stage: {BLIP}~\cite{li2022blip}, {RaSa}~\cite{bai2023rasa} and {IRRA}~\cite{jiang2023irra}. 
We use BLIP as the default choice for generation and retrieval in the following experiments, whose model architecture has been discussed in Section~\ref{section:GTR}.
\textit{Further details regarding the other backbones can be found in the the Appendix.}

\subsection{Ablation Study}

\subsubsection{Effectiveness of Each Component.}
\hfill \\
In this section, we analyze the effectiveness of each key component in the proposed framework, including the basic generation-then-retrieval solution (Basic-GTR), PC-Mask and NP-Train, via a series of ablation studies on three commonly-used datasets with 1\% labeled data. The experimental results are shown in Table~\ref{tab:Effectiveness-of-Each-Component}.\\
\noindent \textbf{Effectiveness of Basic-GTR.}
 Compared to the baseline that solely includes a retrieval stage by training the retrieval model on $1\%$ labeled dataset, the proposed Basic-GTR (M1) achieves substantial improvements across all three datasets (e.g., from $54.91\%$ to $61.94\%$ at Rank-1 on CUHK-PEDES).
 This improvement highlights the effectiveness of Basic-GTR in effectively utilizing unlabeled data for semi-supervised TBPS. 

\noindent \textbf{Effectiveness of PC-Mask.}
The experimental comparison between M1 and M4 provides compelling evidence for the efficacy of PC-Mask. 
Specifically, when incorporating PC-Mask into the Basic-GTR framework, the Rank-1 accuracy is improved by $1.41\%$, $1.41\%$, and $1.05\%$ on the three datasets, respectively. These results confirm that PC-Mask serves as an effective regularization technique during the training process, reducing the interference of noise originating from the pseudo-labeled data.

\noindent \textbf{Effectiveness of NP-Train.}
Beyond applying PC-Mask to mitigate the negative impact of noise, we introduce NP-Train to facilitate noise-robust learning from the perspective of the training strategy. 
\begin{table}
\setlength{\abovecaptionskip}{0.1cm}
    \centering
    \caption{Comparison with different noise measurers on CUHK-PEDES with 1\% labeled data. The default setting is marked in bold.}
    \begin{tabular}{c|ccc|c}
    \toprule
    \textbf{Methods} & \textbf{R-1} & \textbf{R-5} & \textbf{R-10} & \textbf{mAP} \\
    \midrule
    Random sampling & 62.98 & 81.64 & 87.43 & 56.10 \\
    Sentence length & 61.61 & 80.72 & 86.58 & 55.25 \\
    CLIPScore & 62.35 & 81.09 & 87.33 & 56.17 \\
    \textbf{BLIPscore} & \textbf{63.87} & \textbf{82.20} & \textbf{87.70} & \textbf{57.18} \\
    \bottomrule
    \end{tabular}
    \label{tab:Comparison-between-Different-Noise-Measurers}
   \vspace{-0.3cm}
\end{table}
The effectiveness of NP-Train is revealed via the experimental results of M5 \emph{vs.} M1. In particular, when integrating with NP-Train, the performance exhibits significant improvement, with the Rank-1 accuracy increasing by $1.12\%$ on the CUHK-PEDES dataset. These results demonstrate that NP-Train effectively enables the model to learn reliable cross-modal relations in noisy circumstances.

Finally, with the synergy of each component, the proposed framework achieves the promising performance of $63.87\%$, $46.46\%$ and $56.45\%$ at Rank-1 on three datasets using only $1\%$ labeled data and remaining unlabeled data, beating the baseline by a large margin.

\subsubsection{Analysis on the PC-Mask}
\hfill \\
To investigate the inner mechanism of PC-Mask, we provide experimental analysis on the impact of each component inside PC-Mask, including patch-level masking (P-Mask) and channel-level masking (C-Mask).
(1) P-Mask decouples the noisy correspondence by masking in the high-level semantic space. Its effectiveness is demonstrated through the experimental results of M2 \emph{vs.} M1 in Table~\ref{tab:Effectiveness-of-Each-Component}. Integrating P-Mask into the Basic-GTR framework alone leads to significant improvements.
(2) C-Mask exerts the masking on the low-level representation space. 
As shown in the experiments of M3 \emph{vs.} M1, merely adding C-Mask brings consistent improvements at the Rank-1 accuracy on the three datasets by $0.54\%$, $1.03\%$, $0.65\%$, respectively, demonstrating its effectiveness in noise suppression. 
(3) When combining the two masking strategies together, the performance gains further improvement (M4 \emph{vs.} M2/M3), which verifies their underlying complementary nature.

\subsubsection{Analysis on the NP-Train}
\hfill \\
To further study the effect of Noise-guided Progressive Training (NP-Train), we conduct ablation studies on the two key components of NP-Train, namely noise measurer and training scheduler. \\
\noindent \textbf{Comparisons with different noise measurers.}
In NP-Train, the noise measurer is critical in deciding the training curriculum. To investigate the effect of our proposed noise measurer, we compare it with different variants, as shown in Table~\ref{tab:Comparison-between-Different-Noise-Measurers}. 
First, we evaluate the performance using \emph{Random sampling}, where the samples are randomly scheduled without considering noise assessment.
Second, we conduct experiments using a mono-modal noise measurer called \emph{Sentence length}, which determines the noise criterion based solely on the number of words in a text, following the method~\cite{platanios2019competence}. 
In addition, we compare two noise measurers based on cross-modal semantic similarity: \emph{CLIPScore}~\cite{hessel2021clipscore} and \emph{BLIPscore}. 
\emph{CLIPScore} calculates image-text compatibility based on the pretrained CLIP model~\cite{radford2021learning}. 
On the other hand, \emph{BLIPscore} measures cross-modal similarity using our finetuned backbone model, BLIP, trained on a combination of limited labeled data and pseudo-labeled data.
From Table~\ref{tab:Comparison-between-Different-Noise-Measurers}, we can observe that:
(1)	Compared with \emph{Random sampling}, our utilization of \emph{BLIPscore} leads to significant improvements at Rank-1 and mAP by $0.89\%$ and $1.08\%$, respectively. This outcome serves as evidence showcasing the effectiveness of our proposed noise measurer in establishing a meaningful learning curriculum.
(2)	Generally, the application of noise measurers based on cross-modal semantic similarity (last two rows in Table~\ref{tab:Comparison-between-Different-Noise-Measurers}) yields better performance compared to the usage of the mono-modal noise measurer, \emph{Sentence length}. This observation verifies that the noise presented in pseudo-text primarily manifests at the semantic level. Consequently, measuring noise based on cross-modal consistency proves to be more adequate and superior in capturing and quantifying such noise.
(3)	Compared with \emph{CLIPScore}, \emph{BLIPscore} shows better performance (e.g. $1.52\%$ improvement at Rank-1), which demonstrates that vision-language models pretrained on the generic domain are inadequate in accurately measuring fine-grained semantic similarity.

\begin{table}
\setlength{\abovecaptionskip}{0.1cm}
    \centering
    \caption{Comparison with different training schedulers on CUHK-PEDES with 1\% labeled data.}
    \begin{tabular}{c|ccc|c}
    \toprule
    \textbf{Methods} & \textbf{R-1} & \textbf{R-5} & \textbf{R-10} & \textbf{mAP} \\
    \midrule
    Baby step & 63.69 & 82.10 & 87.82 & 57.32 \\
    Root-2 & 62.98 & 81.76 & 87.49 & 56.80 \\
    \textbf{Linear} & \textbf{63.87} & \textbf{82.20} & \textbf{87.70} & \textbf{57.18} \\
    \bottomrule
    \end{tabular}
    \label{tab:Comparison-between-Different-Training-Schedulers}
   \vspace{-1.1cm}
\end{table}

\noindent \textbf{Comparisons with different training schedulers.}
Based on the sorted data from the noise measurer, training scheduler further decides the sequence of training data subsets throughout the training process. To investigate the influence of applying different scheduling strategies, we experiment with another two popular training schedulers including \emph{Baby step}~\cite{bengio2009curriculumLearning, Spitkovsky2010babystep} and \emph{Root-2}~\cite{platanios2019competence}. 
In the \emph{Baby step} scheduler, the sorted training data is initially divided into buckets. Subsequently, the data is incrementally added bucket by bucket, with an increasing noise amount, after a fixed number of epochs. In contrast to our \emph{Linear} scheduler, which adjusts the training data at each epoch, \emph{Baby step} operates as a discrete scheduler, performing scheduling after a fixed number of epochs.
Similar to \emph{Linear}, \emph{Root-2} is also a kind of continuous scheduler which arranges the number of training data at each epoch following a root function. 
The experimental result is shown in Table~\ref{tab:Comparison-between-Different-Training-Schedulers}.
(1) \emph{Linear} performs slightly better in terms of Rank-1 accuracy than \emph{Baby step}. We conjecture that this could be attributed to the fact that \emph{Linear} follows a gentler and smoother way in scheduling the training data compared to \emph{Baby step}, which might be advantageous for the optimization process.
(2) \emph{Linear} surpasses \emph{Root-2} by a significant margin, with a substantial improvement of $0.89\%$ at Rank-1. We propose that this discrepancy stems from the fact that \emph{Root-2} allocates more training time to data with a higher level of noise compared to \emph{Linear}. Consequently, this allocation may inadvertently amplify the impact of noise interference during the optimization process.

\subsection{Comparison with the State-of-the-Arts}
We present the comparison results on CUHK-PEDES, ICFG-PEDES, and RSTPReid in Tables~\ref{tab:Comparison-with-SOTA-methods-on-CUHK-PEDES}-\ref{tab:Comparison-with-SOTA-methods-on-RSTPReid}. 
\emph{The comparison on the recently released dataset UFine6926 is provided in the Appendix.}

Since this paper is the first in semi-supervised TBPS, there are no published works with the same setting that can be fairly compared with the proposed framework.
We thus firstly compare the proposed framework with current published methods, which are usually under fully-supervised learning, to indirectly assess its performance.
Compared with the methods under the fully-supervised setting ($100\%$ Labeled), the proposed framework ($20\%$, $5\%$ or $1\%$ Labeled) exhibits an expected performance gap due to the limited amount of labeled data. However, it still achieves promising results. With only $20\%$ labeled data, the proposed framework achieves competitive results with IRRA, such as $72.86\%$ \emph{vs}. $73.38\%$ in terms of R-1 on CUHK-PEDES. Notably, it even surpasses the strong method RaSa on RSTPReid, \emph{e.g.}, $67.10\%$ \emph{vs}. $66.90\%$. Moreover, the cost savings associated with labeling expenses provided by the proposed framework make it more advantageous for practical applications.

Next, we compare with the unsupervised method GTR (completely abandoning annotations, $0\%$ Labeled) to verify the advantage of the proposed framework in terms of the trade-off between annotation cost and performance. Compared to GTR, the proposed framework demonstrates a significant performance advantage, with respective increases of $16.34\%$, $18.21\%$, and $10.85\%$ at Rank-1 on the three datasets using only $1\%$ labeled data.

Finally, we directly assess the performance of the proposed framework by comparing it with two state-of-the-art methods, IRRA and RaSa, under the same settings\footnote{The results of IRRA and RaSa under the semi-supervised learning setting are obtained by running their published codes with the limited amount of labeled data.}. As observed, the proposed framework outperforms IRRA and RaSa by a large margin across all labeled ratios. Particularly in the most realistic and challenging scenario where only $1\%$ labeled data is available, the proposed framework surpasses IRRA and RaSa by $27.35\%$ and $16.42\%$ at R-1 on CUHK-PEDES, clearly highlighting the unique advantages of the proposed framework.

\begin{table}
\setlength{\abovecaptionskip}{-0.01cm}
\centering
\caption{Comparison with SOTA methods on CUHK-PEDES. ``Labeled'' refers to the amount of labeled training data.}
\resizebox{0.99\columnwidth}{!}{
\begin{tabular}{c|c|c|ccc|c}
\toprule
\textbf{Labeled} & \textbf{Methods} & \textbf{References}  & \makebox[0.045\textwidth][c]{\textbf{R-1}}  & \makebox[0.045\textwidth][c]{\textbf{R-5}} & \makebox[0.05\textwidth][c]{\textbf{R-10}} & \makebox[0.03\textwidth][c]{\textbf{mAP}}\\
\midrule
\multirow{9}{*}{100\%} & ViTAA~\cite{wang2020vitaa} & ECCV 2020 & 55.97 & 75.84 & 83.52 & 51.60  \\
 & DSSL~\cite{zhu2021dssl} & ACMMM-2021 & 59.98 & 80.41 & 87.56 & -  \\
 & SAF~\cite{li2022learning}  & ICASSP-2022 & 64.13 & 82.62 & 88.40 & 58.61 \\
 & LGUR~\cite{shao2022learning} & ACMMM-2022 & 65.25 & 83.12 & 89.00 & -     \\
 & CFine~\cite{yan2022clip}  & TIP-2023 & 69.57 & 85.93 & 91.15 & - \\
 & IRRA~\cite{jiang2023irra}   & CVPR-2023 & 73.38 & 89.93 & 93.71 & 66.13 \\
 & TBPS-CLIP~\cite{cao2023empirical} & AAAI-2024 & 73.54 & 88.19 & 92.35 & 65.38 \\
 & APTM~\cite{yang2023aptm}  & ACMMM-2023 & 76.53 & 90.04 & 94.15 & 66.91 \\
 & RaSa~\cite{bai2023rasa} & IJCAI-2023 & 76.51 & 90.29 & 94.25 & 69.38  \\
\hline
0\%  & GTR~\cite{bai2023text}  & ACMMM-2023 & 47.53 & 68.23 & 75.91 & 42.91 \\
\midrule
\multirow{3}{*}{20\%} & IRRA~\cite{jiang2023irra} & CVPR-2023 & 61.73 & 82.31 & 88.78 & 54.86 \\
& RaSa~\cite{bai2023rasa} & IJCAI-2023 & 67.22 & 85.17 & 90.30 & 58.83 \\
\rowcolor[gray]{.9} & Ours &  - &  72.86 &  88.08 &  92.30 &  65.25 \\
\hline
\multirow{3}{*}{5\%} & IRRA~\cite{jiang2023irra} & CVPR-2023 & 46.91 & 69.61 & 78.25 & 41.63 \\
& RaSa~\cite{bai2023rasa} & IJCAI-2023 & 56.92 & 77.63 & 84.37 & 49.08  \\
\rowcolor[gray]{.9} & Ours & - & 68.76 & 85.15 &  90.56 &  61.82 \\
\hline
\multirow{3}{*}{1\%} & IRRA~\cite{jiang2023irra} & CVPR-2023 & 36.52 & 59.67 & 69.53 & 33.46 \\
& RaSa~\cite{bai2023rasa} & IJCAI-2023 & 47.45 & 67.38 & 74.69 & 39.13 \\
\rowcolor[gray]{.9} & Ours & - &  63.87 & 82.20 & 87.70 &  57.18\\
\bottomrule
\end{tabular}
}
\label{tab:Comparison-with-SOTA-methods-on-CUHK-PEDES}
\vspace{-0.5cm}
\end{table}

\subsection{Extended Experiments and Visualization}
We carry out extended experiments to validate the scalability of the proposed solution using various backbones and evaluate the generalizability of the noise-robust retrieval framework across different ratios of labeled data.
Besides, we also conduct parameter analysis on different masking ratio parameters to investigate the mechanism of our framework.
The experimental results can be found in the \textit{Appendix}. 
For qualitative analysis, we provide retrieval visualization in the \textit{Appendix}, vividly demonstrating the outstanding retrieval capability of the proposed framework.

\begin{table}
\setlength{\abovecaptionskip}{-0.01cm}
\centering
\caption{Comparison with SOTA methods on ICFG-PEDES.}
\resizebox{0.99\columnwidth}{!}{
\begin{tabular}{c|c|c|ccc|c}
\toprule
\textbf{Labeled} & \textbf{Methods} & \textbf{References} & \makebox[0.045\textwidth][c]{\textbf{R-1}}  & \makebox[0.045\textwidth][c]{\textbf{R-5}} & \makebox[0.05\textwidth][c]{\textbf{R-10}} & \makebox[0.03\textwidth][c]{\textbf{mAP}}   \\
\midrule
\multirow{7}{*}{100\%} & Dual Path~\cite{zheng2020dual} & TOMM-2020 & 38.99 &59.44 & 68.41  & -  \\
 & LGUR~\cite{shao2022learning} & ACMMM-2022 & 57.42 &74.97 & 81.45 &-  \\
 & CFine~\cite{yan2022clip} & TIP-2023 & 60.83 & 76.55 & 82.42 & - \\
 & IRRA~\cite{jiang2023irra} & CVPR-2023 & 63.46 & 80.25 & 85.82 & 38.06  \\
 & TBPS-CLIP~\cite{cao2023empirical} & AAAI-2024 & 65.05 & 80.34 & 85.47 & 39.83 \\
 & RaSa~\cite{bai2023rasa} & IJCAI-2023 & 65.28 & 80.40 & 85.12 & 41.29 \\
 & APTM~\cite{yang2023aptm} & ACMMM-2023 & 68.51 & 82.99	& 87.56 & 41.22 \\
 \hline
0\% & GTR~\cite{bai2023text} & ACMMM-2023 & 28.25 & 45.21 & 53.51 & 13.82 \\
\midrule
\multirow{3}{*}{20\%} & IRRA~\cite{jiang2023irra} & CVPR-2023 & 50.32 & 70.34 & 77.83 & 27.83 \\
& RaSa~\cite{bai2023rasa} & IJCAI-2023 & 54.55 & 71.51 & 77.34 & 27.99 \\
\rowcolor[gray]{.9} & Ours & - & 61.65 & 77.06 & 82.66 &  38.53 \\
\hline
\multirow{3}{*}{5\%} & IRRA~\cite{jiang2023irra} & CVPR-2023 & 35.61 & 56.95 & 66.00 & 19.04 \\
& RaSa~\cite{bai2023rasa} & IJCAI-2023 & 40.88 & 58.36 & 65.42 & 15.86 \\
\rowcolor[gray]{.9} & Ours & - & 55.96 &  72.12 &  78.31 & 34.12 \\
\hline
\multirow{3}{*}{1\%} & IRRA~\cite{jiang2023irra} & CVPR-2023 & 22.72 & 41.12 & 50.28 & 11.79 \\
& RaSa~\cite{bai2023rasa} & IJCAI-2023 & 30.28 & 47.64 & 55.11 & 9.80 \\
\rowcolor[gray]{.9} & Ours & - & 46.46  & 64.34 & 71.60 & 26.90 \\
\bottomrule
\end{tabular}
}
\vspace{-0.3cm}
\label{tab:Comparison-with-SOTA-methods-on-ICFG-PEDES}
\end{table}

\begin{table}
\setlength{\abovecaptionskip}{-0.01cm}
\centering
\caption{Comparison with SOTA methods on RSTPReid.}
\resizebox{0.99\columnwidth}{!}{
\begin{tabular}{c|c|c|ccc|c}
\toprule
\textbf{Labeled} & \textbf{Methods} & \textbf{References}  & \makebox[0.045\textwidth][c]{\textbf{R-1}}  & \makebox[0.045\textwidth][c]{\textbf{R-5}} & \makebox[0.05\textwidth][c]{\textbf{R-10}} & \makebox[0.03\textwidth][c]{\textbf{mAP}}   \\
\midrule
\multirow{7}{*}{100\%} & DSSL~\cite{zhu2021dssl} & ACMMM-2021 & 39.05 & 62.60 & 73.95 & - \\
 & SSAN~\cite{ding2021semantically} & Arxiv-2021 & 43.50 & 67.80 & 77.15 & - \\
 & CFine~\cite{yan2022clip} & TIP-2023 & 50.55 & 72.50 & 81.60 & - \\
 & IRRA~\cite{jiang2023irra} & CVPR-2023 & 60.20 & 81.30 & 88.20 & 47.17 \\
 & TBPS-CLIP~\cite{cao2023empirical}  & AAAI-2024 & 61.95 & 83.55 & 88.75 & 48.26 \\
 & RaSa~\cite{bai2023rasa}  & IJCAI-2023 & 66.90 & 86.50 & 91.35 & 52.31 \\
 & APTM~\cite{yang2023aptm}  & ACMMM-2023 & 67.50 & 85.70 & 91.45 & 52.56 \\
 \hline
0\% & GTR~\cite{bai2023text} & ACMMM-2023 & 45.60 & 70.35 & 79.95 & 33.30 \\
\midrule
\multirow{3}{*}{20\%} & IRRA~\cite{jiang2023irra} & CVPR-2023 & 50.85 & 74.60 & 84.70 & 39.19 \\
& RaSa~\cite{bai2023rasa} & IJCAI-2023 & 58.80 & 80.90 & 87.75 & 44.66 \\
\rowcolor[gray]{.9} & Ours & - & 67.10 & 85.85 & 92.00 & 52.84\\
\hline
\multirow{3}{*}{5\%} & IRRA~\cite{jiang2023irra} & CVPR-2023 & 53.45 & 77.05 & 86.40 & 40.18 \\
& RaSa~\cite{bai2023rasa} & IJCAI-2023 & 59.35 & 80.65 & 88.00 & 44.22 \\
\rowcolor[gray]{.9} & Ours & - & 65.60 & 84.75 & 90.75 & 52.29 \\
\hline
\multirow{3}{*}{1\%} & IRRA~\cite{jiang2023irra} & CVPR-2023 & 32.25 & 58.85 & 70.10 & 26.31 \\
& RaSa~\cite{bai2023rasa} & IJCAI-2023 & 46.35 & 69.90 & 78.55 & 32.58 \\
\rowcolor[gray]{.9} & Ours & - & 56.45 & 78.95 & 87.05 & 44.45 \\
\bottomrule
\end{tabular}
}
\vspace{-0.3cm}
\label{tab:Comparison-with-SOTA-methods-on-RSTPReid}
\end{table}

\section{Conclusion}
In this paper, we explore the practical setting of semi-supervised TBPS. We propose a two-stage basic solution based on generation-then-retrieval. Firstly, we address the lack of annotations by generating pseudo-labeled samples. Then, we train a retrieval model in a supervised manner. To handle the noise interference from pseudo-labeled samples during retrieval training, we introduce a noise-robust retrieval framework. This framework consists of two key strategies: Hybrid Patch-Channel Masking (PC-Mask) and Noise-Guided Progressive Training (NP-Train). PC-Mask decouples noisy cross-modal correspondence by masking input data at the patch-level and channel-level to prevent overfitting to noisy supervision. NP-Train enables noise-robust learning by scheduling training progressively based on the noise level of pseudo-labeled samples. Through extensive experiments on multiple TBPS benchmarks, we demonstrate the effectiveness and superiority of our proposed framework in the semi-supervised setting.

\bibliographystyle{ACM-Reference-Format}
\bibliography{base}

\appendix
\section{Experimental Settings}
\subsection{Datasets}
We evaluate the proposed framework based on on three widely-used TBPS datasets: CUHK-PEDES~\cite{li2017person}, ICFG-PEDES~\cite{ding2021semantically} and RSTPReid~\cite{zhu2021dssl}, and a newly released dataset: UFine6926~\cite{zuo2023ufinebench}.\\
\textbf{CUHK-PEDES}~\cite{li2017person} is the first benchmark for TBPS, with $40,206$ images, $80,440$ texts, and $13,003$ identities. Each image has an average of two textual descriptions. Following the official data split, the training dataset contains $34,054$ images and $68,126$ texts from $11,003$ identities, the validation set consists of $3,078$ images and $6,158$ texts from $1,000$ identities, and the test set has $3,074$ images and $6,156$ texts from $1,000$ identities.\\
\textbf{ICFG-PEDES}~\cite{ding2021semantically} has $54,522$ images from $4,102$ identities. Each image has one textual description. The dataset is split into a training set with $34,674$ images from $3,102$ identities, and a test set with $19,848$ images from $1,000$ identities.\\
\textbf{RSTPReid}~\cite{zhu2021dssl} consists of $20,505$ images of $4,101$ identities. Each identity has five images from different cameras, each with two textual descriptions. The training, validation, and test datasets have $3,701$, $200$, and $200$ identities, respectively.\\
\textbf{UFine6926}~\cite{zuo2023ufinebench} contains
$26,206$ images and $52,412$ descriptions of $6,926$ persons totally. The textual annotations have ultra-fine granularity, with the average word count three to four times that of the previous datasets. The training and test set contains $4926$ and $2,000$ identities respectively.

\subsection{Evaluation Metrics}
We adopt the Rank-k (R-k for simplicity, k=$1, 5, 10$) as the evaluation metrics. These metrics assess the probability of locating at least one matching person image within the top-k candidates, given a text as the query. Also, we adopt the mean average precision (mAP) as a complementary metric, which measures the average retrieval performance in scenarios where multiple ground-truths exist.

\subsection{Implementation Details}
We conduct all experiments on $4$ NVIDIA A$40$ GPUs.
For the generation model, we finetune it with a learning rate of $1e\!-\!5$ and a batch size of $64$ for $10$ epochs, utilizing the Adam optimizer~\cite{kingma2014adam}. The retrieval model, on the other hand, is trained for $20$ epochs with a learning rate of $1e\!-\!5$ and a batch size of $64$. We applied random horizontal flipping as a form of data augmentation for the image data.
In the case of Hybrid Patch-channel Masking, we set the patch-level masking rates as $\rho ^v = 0.2$ and $\rho ^t = 0.1$, respectively, the channel-level masking rates as $\beta ^v = 0.1$ and $\beta ^t = 0.1$.

\subsection{Backbones}
The proposed solution offers flexibility in selecting image captioning and text-based person retrieval backbones. Existing methods can be readily substituted as backbones. In our experiments, we employ multiple backbones, with BLIP being the default choice for generation and retrieval. The model architecture of BLIP has been discussed in the main paper.

\noindent \textbf{Generation.}
Apart from \textbf{BLIP}, the \textbf{BLIP-2}~\cite{li2023blip2} is also utilized for image captioning. BLIP-2 is a versatile and highly efficient framework that leverages off-the-shelf frozen image encoders and large language models via a lightweight querying transformer. It enables effective vision-language pretraining and yields impressive results across various vision-language tasks, including image captioning.

\noindent \textbf{Retrieval.}
In addition to \textbf{BLIP}, we also conduct experiments on the recent popular TBPS methods \textbf{RaSa}~\cite{bai2023rasa} and \textbf{IRRA}~\cite{jiang2023irra}. RaSa leverages relation-aware learning and sensitivity-aware learning to mitigate the impact of weak cross-modal correspondence and facilitate representation learning in TBPS. IRRA introduces a cross-modal implicit relation reasoning module to enhance global image-text matching and a similarity distribution matching loss to amplify the correlations between matching pairs.

\section{Extended Experiments}
\subsection{Study on Scalability of Two-stage Solution}
We propose a scalable solution for semi-supervised TBPS, in which the generation and retrieval models can be flexibly changed with existing methods. To verify the scalability of the proposed solution, in addition to the default backbone BLIP~\cite{li2022blip}, we conduct a series of experiments with different model variants, including BLIP-2~\cite{li2023blip2} and BLIP-2~\cite{li2023blip2} for generation, RaSa~\cite{bai2023rasa} and IRRA~\cite{jiang2023irra} for retrieval. The experimental results are shown in Table~\ref{tab:Ablations-of-Scalability}.

\begin{table}
  \setlength{\abovecaptionskip}{-0.01cm} 
    \centering
    \tabcolsep=3.5pt
    \caption{Comparison with different variants of the proposed solution on CUHK-PEDES with 1\% labeled data.}
    \begin{tabular}{c|c|ccc|c}
    \toprule
    \textbf{Generation} & \textbf{Retrieval} & \textbf{R-1} & \textbf{R-5} & \textbf{R-10} & \textbf{mAP} \\
    \midrule
    \multicolumn{6}{c}{\textit{Using different generation models}} \\
    \hline
    \textbf{BLIP} & \textbf{BLIP} & \textbf{63.87} & \textbf{82.20} & \textbf{87.70} & \textbf{57.18}  \\
    BLIP-2 & BLIP & 62.12 & 80.96 & 87.18 & 56.57 \\
    \hline
    \multicolumn{6}{c}{\textit{Using different retrieval models}} \\
    \hline
    \textbf{BLIP} & \textbf{BLIP} & \textbf{63.87} & \textbf{82.20} & \textbf{87.70} & \textbf{57.18}  \\
    BLIP & RaSa & 61.96	& 80.69 & 86.87 & 55.29  \\
    BLIP & IRRA & 56.17	& 76.49 & 83.85 & 50.99  \\
    \bottomrule
    \end{tabular}
        \vspace{-0.4cm}
   \label{tab:Ablations-of-Scalability}
\end{table}

\noindent\textbf{Generation}.
From Table~\ref{tab:Ablations-of-Scalability}, we can observe that the proposed framework, whether utilizing BLIP or BLIP-2, consistently achieves excellent performance, thereby strongly validating its scalability.
Furthermore, it is worth noting that the proposed framework with the default BLIP exhibits slightly stronger performance compared to using BLIP-2.
We hypothesize that this discrepancy can be attributed to their distinct decoding strategies.
BLIP employs nucleus sampling~\cite{Holtzman2020Nucleus} as its decoding strategy, which is a stochastic method, while BLIP-2 utilizes the deterministic decoding method of beam search~\cite{freitag2017beam}.
Due to its stochastic nature, nucleus sampling in BLIP tends to generate a greater diversity of pseudo-texts~\cite{li2022blip} when compared to beam search in BLIP-2.
This diversity characteristic aligns well with the open-form nature of textual queries in TBPS, thereby benefiting the subsequent training of the retrieval model.

\noindent\textbf{Retrieval}.
From Table~\ref{tab:Ablations-of-Scalability}, we can clearly see that the proposed framework, whether using BLIP or RaSa, yields commendable results.
In particular, the proposed framework with BLIP surpasses the performance of that with RaSa. We conjecture that it is because BLIP harvests stronger vision-language alignment capability than RaSa.
We also observed that the proposed framework with IRRA achieves a relatively mediocre performance. This could potentially be attributed to IRRA's higher sensitivity to misalignment and noise introduced by the pseudo-texts.

\begin{figure}
  \setlength{\abovecaptionskip}{0.01cm}
  \centering
  \includegraphics[width=\linewidth]{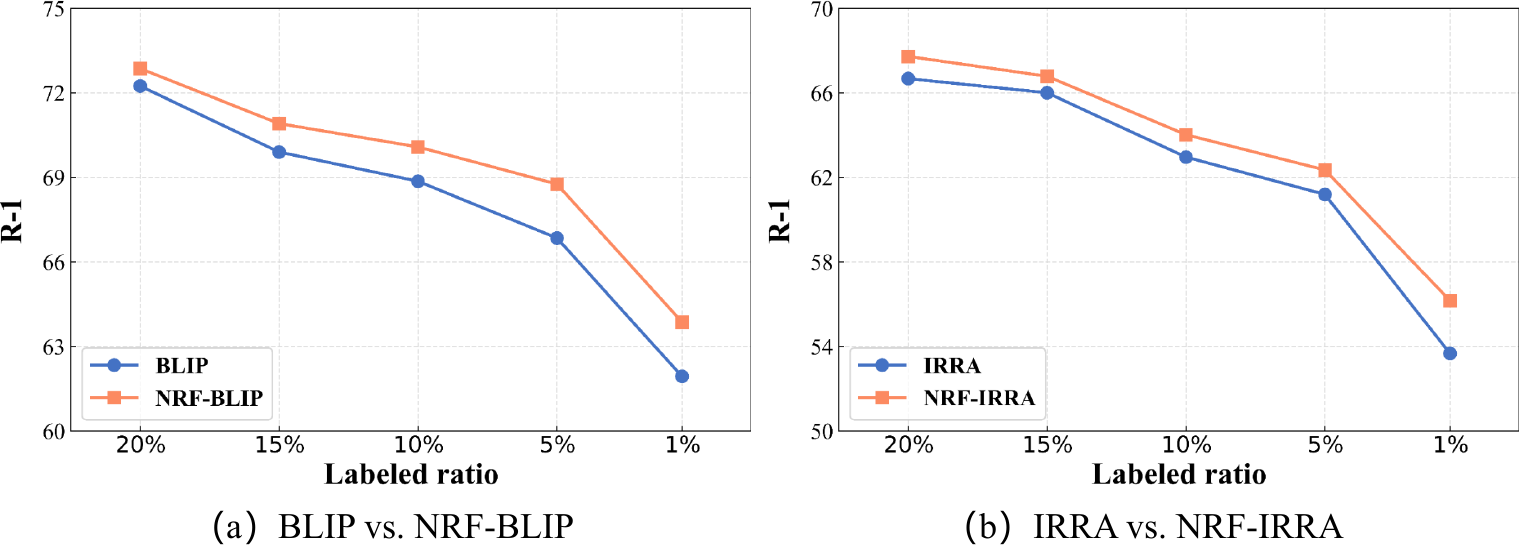}
  \caption{Study on effectiveness and generalizability of the proposed noise-robust retrieval framework (NRF) on CUHK-PEDES with decreasing ratios of labeled data.}
  \label{fig:generalizability-analysis}
  \vspace{-0.7cm}
\end{figure}

\subsection{Study on Effectiveness and Generalizability of Noise-robust Retrieval Framework}
To comprehensively validate the effectiveness of the proposed noise-robust retrieval framework (NRF), we conduct experiments with default retrieval backbone BLIP~\cite{li2022blip} on CUHK-PEDES across different labeled ratios. Since the noise problem in pseudo texts is more severe with limited labeled data, we choose to experiment under a small labeled ratio $20\%$ and further decrease it to $1\%$ with an interval of $5\%$ for exploration of learning with increasing noise. We denote BLIP equipped with NRF as NRF-BLIP. In addition, to verify the generalizability of NRF to other retrieval model, we extend IRRA~\cite{jiang2023irra} by NRF, denoted by NRF-IRRA. The experimental results are shown in Figure.~\ref{fig:generalizability-analysis}, from which we can observe that:
(1) NRF-BLIP consistently outperforms BLIP across all labeled ratios, which strongly verifies the effectiveness of our NRF in learning under the noisy circumstance.
(2) NRF-IRRA is evidently superior to IRRA in all tests, which shows the generalizability of NRF.
(3) With the decrease of labeled ratio, the noise problem in the pseudo-texts gets more severe, which poses more challenges on the subsequent learning of retrieval model. NRF brings more performance gains to the underlying retrieval model under smaller labeled ratios, which demonstrates the robustness of NRF. 

\subsection{Study on the Labeled Ratio}
\begin{figure}
  \setlength{\abovecaptionskip}{0.01cm}
  \centering
  \includegraphics[width=\linewidth]{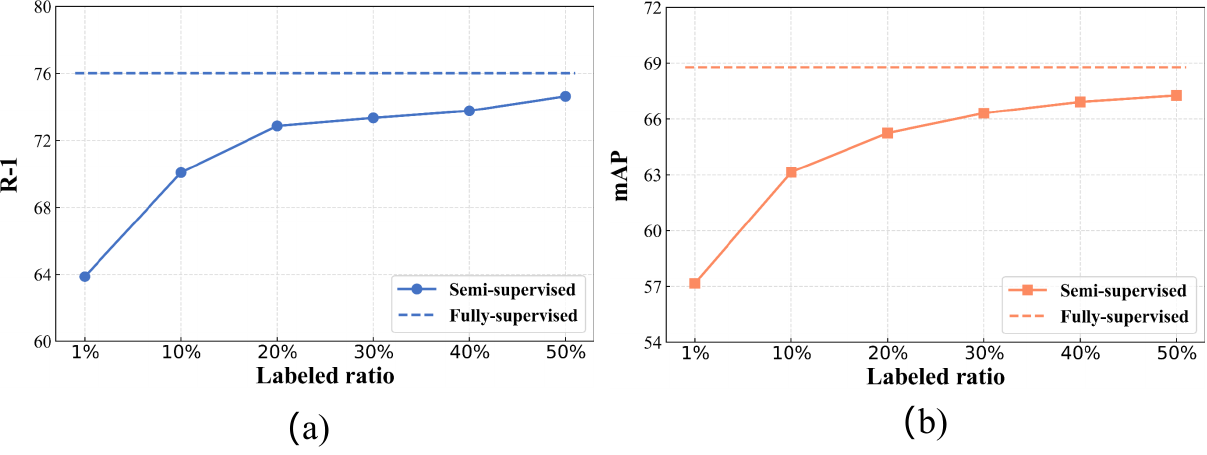}
  \caption{Study on the labeled ratio on CUHK-PEDES.}
  \label{fig:labeled_ratio-analysis}
  \vspace{-0.7cm}
\end{figure}

To research the influence of the labeled ratio in the semi-supervised setting, we conduct experiments on CUHK-PEDES by varying the labeled ratio from $1\%$ to $50\%$ with an interval of $10\%$. We also provide the performance of the default backbone BLIP under the fully-supervised setting as the performance upper bound for reference, where $100\%$ labeled data is used for training. The experimental results are shown in Fig.~\ref{fig:labeled_ratio-analysis}, from which we can observe that the performance of our framework gradually rises with the increasing amount of labeled data. When using only $20\%$ labeled data, our framework can approach the fully-supervised counterpart with a relatively small gap. As the labeled ratio continues to increase, the growth rate of performance gets slower. We conjecture that this is possibly because the generation model has already learned most knowledge of person captioning with $20\%$ labeled data. Further incorporating more labeled data may not increase the quality of generated pseudo-texts substantially. Thus, less performance gains would be obtained by the retrieval model trained subsequently.

\subsection{Parameter Analysis.}
\begin{figure}
\setlength{\abovecaptionskip}{0.01cm}
  \centering
  \includegraphics[width=\linewidth]{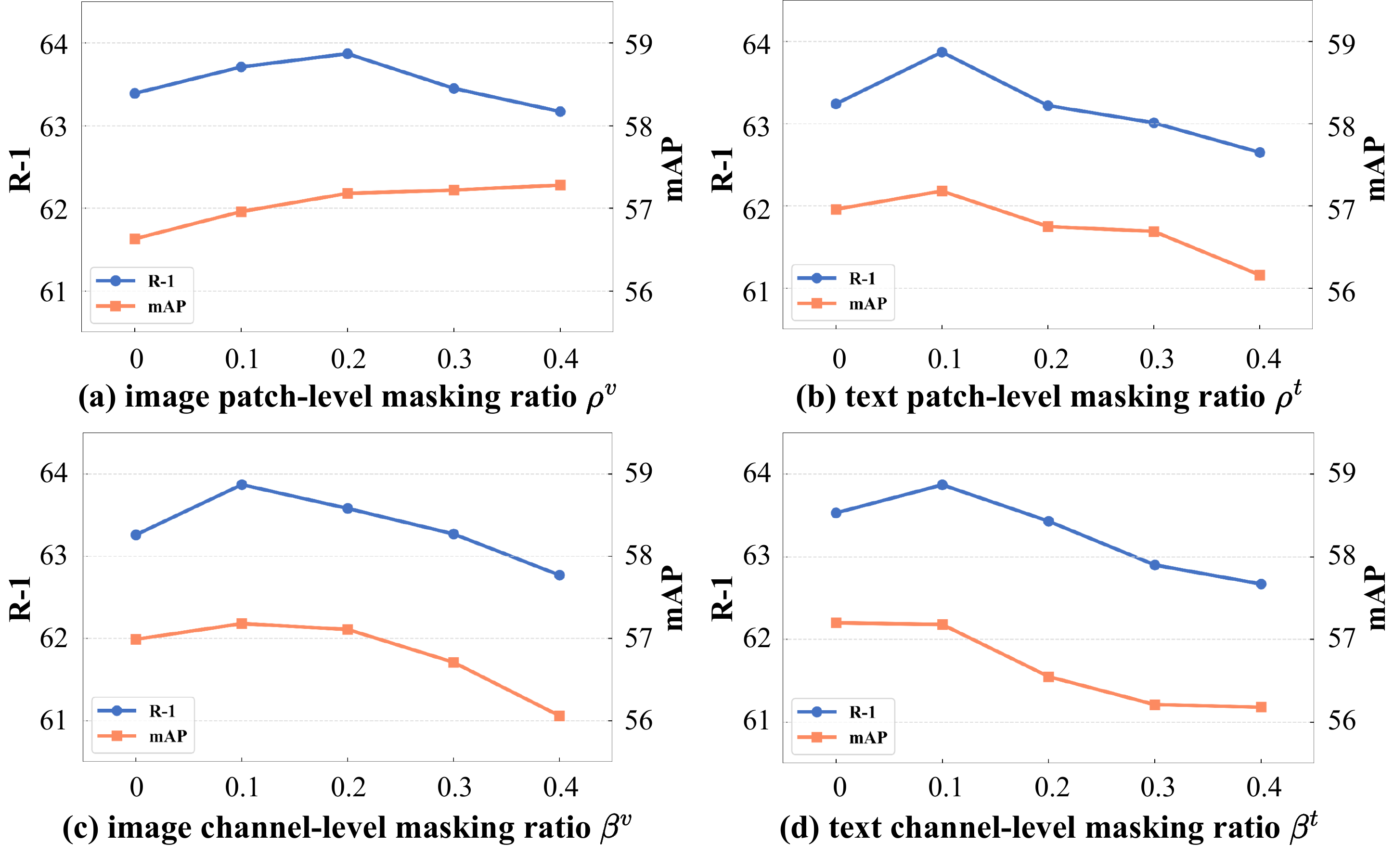}
  \caption{Retrieval performance under varying parameters of PC-Mask on CUHK-PEDES with 1\% labeled data.}
  \label{fig:param-analysis}
\vspace{-0.3cm}
\end{figure}

We introduce the Hybrid Patch-Channel Masking (PC-Mask) strategy to suppress the noise in the retrieval stage, where the parameter $\rho ^v$ and $\rho ^t$ is used to control the ratio of image and text patch-level masking, $\beta ^v$ and $\beta ^t$ is used to control the ratio of image and text channel-level masking, respectively. The influence of these parameters is shown in Fig.~\ref{fig:param-analysis}.
(1) In terms of patch-level masking on both image and text, it can be observed that as $\rho ^v$ or $\rho ^t$ increase, Rank-1 and mAP initially rise and subsequently decline. The early-stage performance gain can be attributed to the increased efficacy of masking in separating noisy correspondences. However, as $\rho ^v$ continues to increase, valuable and essential semantic relations may be disrupted, impeding the model's ability to learn cross-modal alignment and ultimately resulting in decreased performance.
The peak Rank-1 performance is achieved at $\rho ^v=0.2$ and $\rho ^t=0.1$, which are adopted in our experiments.
(2) In terms of channel-level masking on both image and text, the Rank-1 and mAP also exhibit an initial increase followed by a subsequent decrease as the masking ratios increase. Notably, channel-level masking shows a more pronounced decline in performance compared to patch-level masking when using a high masking ratio. We speculate that this is due to channel-level masking directly affecting the low-level feature channels, which can have a more direct and significant impact on representation learning compared to the higher-level patch-level masking.
The highest Rank-1 performance is achieved at $\beta ^v=0.1$ and $\beta ^t=0.1$, which are the adopted ratios in our experiments.

\subsection{Comparison with SoTAs on UFine6926}
We compare the proposed framework with the state-of-the-art methods on a newly released dataset UFine6926~\cite{zuo2023ufinebench}. The textual annotations in UFine6926 show ultra-fine granularity, which poses a new challenge to the research of text-based person search. The experimental results are shown in Table~\ref{tab:Comparison-with-SOTA-methods-on-UFine6926}. First, under the semi-supervised setting, our framework outperforms IRRA~\cite{jiang2023irra} and RaSa~\cite{bai2023rasa} consistently across all labeled ratios. In particular, when only $1\%$ labeled data is available, the proposed framework surpasses IRRA and RaSa by $15\%$ and $19.11\%$ in terms of R-1, highlighting the advantages of the proposed framework. Second, when using $20\%$ labeled data, our framework performs favorably against many SoTA methods from the fully-supervised setting. For example, our framework outperforms LGUR~\cite{shao2022learning} and SSAN~\cite{ding2021semantically} by $10.29\%$ and $5.89\%$ at R-1, respectively. The promising performance on this ultra fine-grained dataset demonstrates the generalization ability and fitness of our framework towards the complex scenarios. 

\begin{table}
\centering
\caption{Comparison with SOTA methods on UFine6926. ``Labeled'' refers to the amount of labeled training data.}
\resizebox{0.99\columnwidth}{!}{
\begin{tabular}{c|c|c|ccc|c}
\toprule
\textbf{Labeled} & \textbf{Methods} & \textbf{References}  & \makebox[0.045\textwidth][c]{\textbf{R-1}}  & \makebox[0.045\textwidth][c]{\textbf{R-5}} & \makebox[0.05\textwidth][c]{\textbf{R-10}} & \makebox[0.03\textwidth][c]{\textbf{mAP}}\\
\midrule
\multirow{5}{*}{100\%} 
& NAFS~\cite{gao2021contextual} & Arxiv-2021 & 64.11 & 80.32 & 85.05 & 63.47 \\
 & LGUR~\cite{shao2022learning} & ACMMM-2022 & 70.69 & 84.57 & 89.91 & 68.93     \\
& SSAN~\cite{ding2021semantically} & Arxiv-2021 & 75.09 & 88.63 & 92.84 & 73.14     \\
 & IRRA~\cite{jiang2023irra}   & CVPR-2023 & 83.53 & 92.94 & 95.95 & 82.79 \\
 & CFAM(B/16)~\cite{zuo2023ufinebench}   & CVPR-2024 & 85.55 & 94.51 & 97.02 & 84.23 \\
\midrule
\multirow{3}{*}{20\%} & IRRA~\cite{jiang2023irra} & CVPR-2023 & 71.80 & 86.49 & 91.75 & 70.18 \\
& RaSa~\cite{bai2023rasa} & IJCAI-2023 & 76.95 & 89.66 & 93.36 & 73.16 \\
\rowcolor[gray]{.9} & Ours &  - & 80.98 & 92.25 & 95.29 & 79.40 \\
\hline
\multirow{3}{*}{5\%} & IRRA~\cite{jiang2023irra} & CVPR-2023 & 54.71 & 74.88 & 82.80 & 54.35 \\
& RaSa~\cite{bai2023rasa} & IJCAI-2023 & 58.10 & 75.86 & 82.05 & 53.84  \\
\rowcolor[gray]{.9} & Ours & - & 75.60 & 88.75 & 92.83 & 74.06 \\
\hline
\multirow{3}{*}{1\%} & IRRA~\cite{jiang2023irra} & CVPR-2023 & 38.04 & 58.84 & 68.96 & 38.52 \\
& RaSa~\cite{bai2023rasa} & IJCAI-2023 & 33.93 & 52.32 & 60.46 & 31.01 \\
\rowcolor[gray]{.9} & Ours & - &53.04 & 71.24 & 78.98 & 52.58 \\
\bottomrule
\end{tabular}
}
\label{tab:Comparison-with-SOTA-methods-on-UFine6926}
\vspace{-0.3cm}
\end{table}

\section{Visualization Analysis}
\begin{figure}
\setlength{\abovecaptionskip}{0.1cm}
  \centering
  \includegraphics[width=\linewidth]{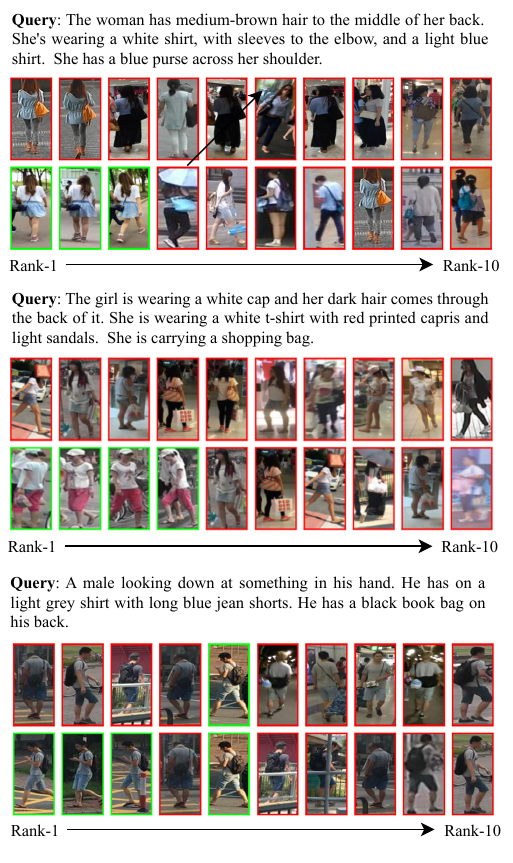}
  \caption{Visualization of top-10 retrieval results on CUHK-PEDES. The first row in each example presents the retrieval results from the baseline, and the second row shows the results from our framework.
Correct/Incorrect images are marked by green/red rectangles.}
  \label{fig:visualization-retrieval-result}
\vspace{-0.3cm}
\end{figure}

\noindent \textbf{Visualization of Retrieval Results.}
We exhibit three top-10 retrieval examples of the baseline and our framework trained with $1\%$ labeled data in Fig.~\ref{fig:visualization-retrieval-result}, where the first row and the second row in each example present the retrieval
results from baseline and our framework, respectively. It can be seen
that our framework can retrieve the corresponding pedestrian images
for a query text more accurately. 
This is mainly due to the
capability of our two-stage solution in exploiting both labeled and unlabeled data and the alleviation of noise interference by the powerful noise-robust retrieval framework. The visualization
vividly demonstrates the effectiveness of our framework.

\begin{figure}
\setlength{\abovecaptionskip}{0.2cm}
  \centering
  \includegraphics[width=\linewidth]{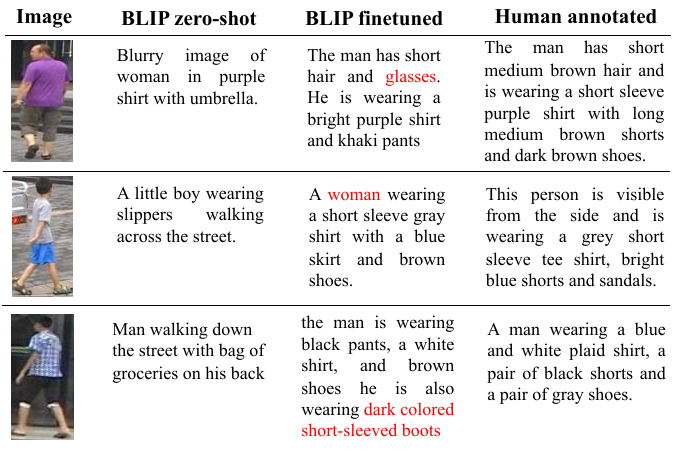}
  \caption{More visualization results of human annotated texts and generated pseudo-texts from the vision-language model BLIP~\cite{li2022blip}
under the zero-shot setting and finetuned on $1\%$ labeled data.
Pseudo-texts from zero-shot BLIP tend to be coarse-grained
while those from finetuned BLIP possess more fine-grained
details but may contain inevitable noise. The noise is highlighted in red.}
  \label{fig:more-example-noise}
\vspace{-0.3cm}
\end{figure}

\noindent \textbf{More Examples of the Generated Pseudo-texts.} 
We provide more visualization results of human annotated texts and
generated pseudo-texts from the vision-language model
BLIP~\cite{li2022blip} under the zero-shot setting and finetuned on $1\%$ labeled data in Fig.~\ref{fig:more-example-noise}. We can see that pseudo-texts from zero-shot BLIP tend to be
coarse-grained while those from finetuned BLIP possess
more fine-grained details but may contain inevitable noise, as highlighted in red. Towards the noise problem in pseudo-texts, we propose the noise-robust retrieval framework to particularly address it.

\section{Limitations}
For the novel and resource-friendly semi-supervised TBPS, we develop a two-stage solution based on generation-then-retrieval. Based on it, our primary focus is on addressing the noise problem of pseudo-texts in the retrieval stage, leading to the proposal of a noise-robust retrieval framework.
Furthermore, we acknowledge the importance of improving the performance of the generation model, as it can help mitigate the noise problem in the subsequent retrieval stage. However, this particular aspect remains unexplored in this paper.
Particularly, current advanced generation models are pretrained vision-language models that exhibit exceptional performance in general domains. However, the effective adaptation of these pretrained vision-language models to the person-specific domain, given limited labeled data, poses an important challenge within our proposed two-stage solution.
In the future, we will explore potential directions from the generation perspective. For instance, leveraging the robust few-shot learning capabilities of multimodal large language models could enable the generation of higher-quality pseudo-texts, thereby further facilitating the advancement of semi-supervised TBPS.

\end{document}